\documentclass[11pt]{article}

\usepackage{acl}

\usepackage{times}
\usepackage{latexsym}

\usepackage[T1]{fontenc}

\usepackage[utf8]{inputenc}

\usepackage{microtype}

\usepackage{inconsolata}

\usepackage{graphicx}
\usepackage{amsmath}
\usepackage[ruled,lined,linesnumbered]{algorithm2e}
\SetKw{KwPhase}{}

\usepackage{float}
\usepackage{graphicx}
\usepackage{multirow}
\usepackage{subcaption}
\usepackage{booktabs}
\usepackage{booktabs}
\usepackage{xcolor}
\usepackage{tcolorbox}
\title{THRD: A Training-Free Multi-Turn Defense Framework for Jailbreak Attacks on Large Language Models}

\renewcommand{\thefootnote}{\fnsymbol{footnote}}
\author{
  \textbf{Zhiqing Ma\textsuperscript{1}},
  \textbf{Zhonghao Xu\textsuperscript{1}},
  \textbf{Dong Yu\textsuperscript{1}},
  \textbf{Chen Kang\textsuperscript{1}},
\\
  \textbf{Changliang Li\thanks{Corresponding authors}},
  \textbf{Pengyuan Liu\textsuperscript{1}\footnotemark[2]}
\\
\\
  \textsuperscript{1}Beijing Language and Culture University
}

\begin{document}
\maketitle
\renewcommand{\thefootnote}{\arabic{footnote}}
\setcounter{footnote}{0}
\begin{abstract}
Multi-turn jailbreak attacks pose a growing threat to LLMs by exploiting conversational dynamics such as gradual escalation and cross-turn coordination. Existing defenses either rely on costly retraining---often degrading model utility---or apply single-turn analysis independently at each turn, failing to capture how risk accumulates along interaction trajectories. We observe that safety behavior in multi-turn interaction is trajectory-dependent: dialogue history continuously reshapes the model's conditioning context, making it insufficient to evaluate each turn in isolation. Motivated by this insight, we present \textsc{THRD}, the first training-free framework that explicitly models temporal risk accumulation for multi-turn jailbreak defense. \textsc{THRD} integrates four modules: a Turn-level Risk Assessor (TRA) for instantaneous risk estimation, a Historical Context Analyzer (HCA) for cross-turn intent escalation detection, a Response Evaluator (RE) for identifying facilitative outputs, and a Decision Module that combines these signals through a time-evolving scoring mechanism with attenuation-based modulation and trend-aware adjustment. Experiments against state-of-the-art multi-turn attacks---including tree-search-based and multi-agent collaborative methods---across two target models show that \textsc{THRD} reduces ASR to 0.2--4.0\% while preserving model utility within 1.5\% degradation on MMLU and GSM8K. Ablation studies confirm non-redundant module contributions and stable cross-architecture generalization. Analysis of first rejection triggers reveals that over 70\% of multi-turn attacks require Turn~2 or later to detect, validating the necessity of explicit temporal aggregation.
\end{abstract}

\section{Introduction}

\begin{figure*}[t]
    \centering
    \includegraphics[width=0.95\textwidth]{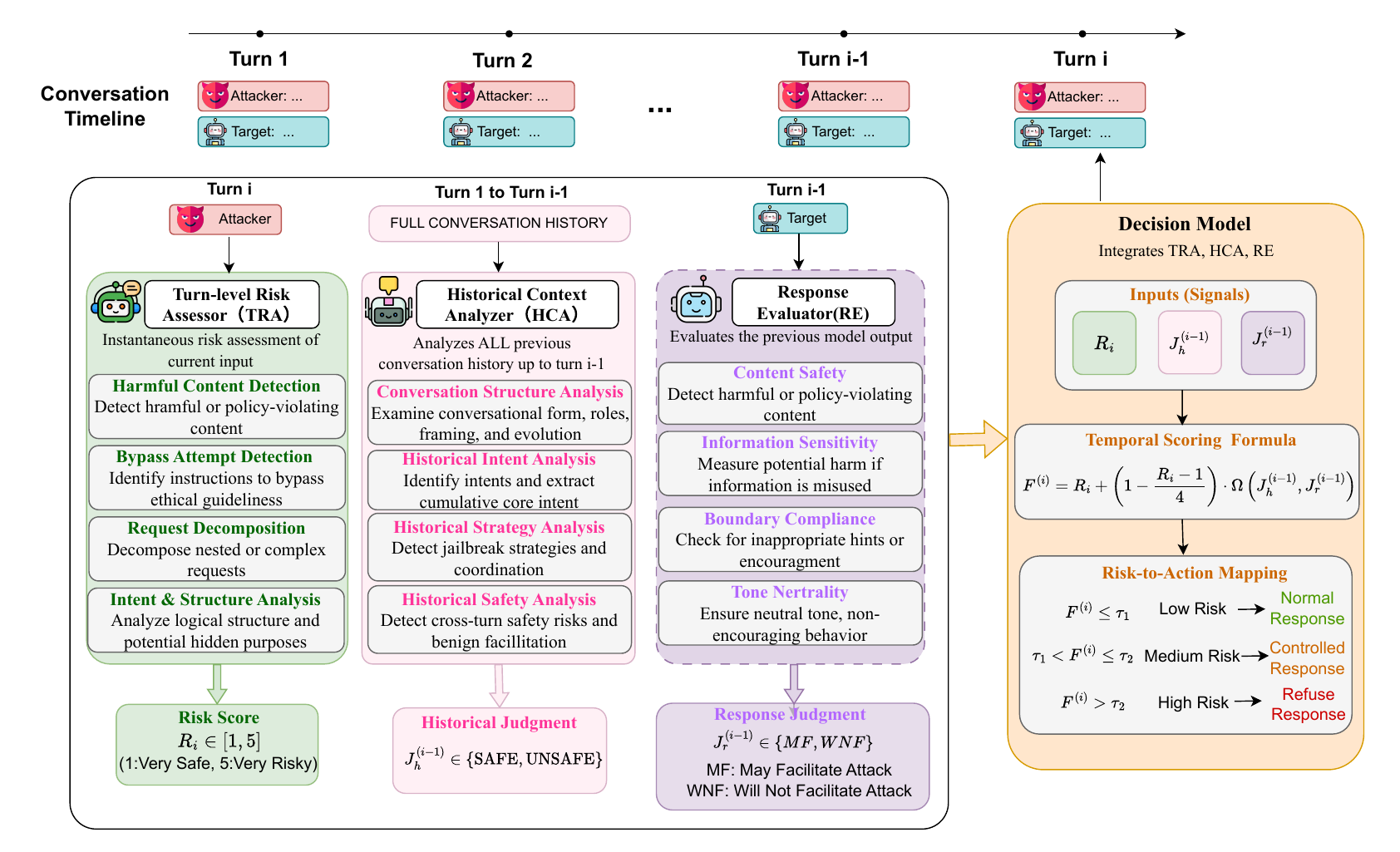}
    \caption{Overview of the THRD framework at turn $i$. TRA evaluates the current input, HCA analyzes conversation history, RE assesses the generated response, and the Decision Module integrates these signals into a time-evolving risk score to determine the final response strategy.}
    \label{fig:framework}
\end{figure*}

Large Language Models (LLMs) such as GPT-5~\cite{openai2025gpt5}, Claude 4~\cite{anthropic2025claude4}, Llama-3~\cite{dubey2024llama}, and Qwen2.5~\cite{qwen2025qwen25technicalreport} have been widely deployed across critical applications. Despite extensive safety alignment efforts through supervised fine-tuning and reinforcement learning from human feedback~\cite{ouyang2022training,christiano2017deep}, these models remain vulnerable to jailbreak attacks.

Early single-turn attacks primarily relied on handcrafted prompts~\cite{liu2023jailbreaking,shen2024anything} or automated optimization methods including GCG~\cite{zou2023universal}, AutoDAN~\cite{ICLR2024_f83cb637}, PAIR~\cite{chao2025jailbreaking}, and GPTFuzzer~\cite{yu2023gptfuzzer}. More recently, multi-turn jailbreak attacks have emerged as a challenging threat by exploiting conversational interactions that gradually weaken safety boundaries across turns~\cite{russinovich2025great,ren2024derail,rahman2025xteaming,zhou2025tempest,ying-etal-2025-reasoning}.

Defending against multi-turn jailbreak attacks remains challenging because harmful intent often emerges gradually across conversation turns rather than through a single explicit request. Attackers can further adapt their strategies based on intermediate model responses, while seemingly benign turns---including model outputs---may collectively steer the interaction toward harmful outcomes. Training-based defense methods~\cite{ding-etal-2025-sdgo,jiang2025metadefense,lin-etal-2024-mitigating,zhai2023investigating} struggle in this setting, as static retraining cannot easily cover the combinatorial and adaptive space of interaction-driven attack trajectories. Existing inference-time defense methods~\cite{ding-etal-2025-act,zhao2025safebehavior,hu2025ccfc}, on the other hand, often rely on single-turn analysis or static rules, limiting their ability to detect attacks whose risk becomes apparent only through cross-turn accumulation and response-level facilitation.

A key limitation of existing defenses is the implicit assumption that safety decisions can be made based on the current input in isolation. In multi-turn interactions, however, dialogue history continuously reshapes the model's conditioning context, meaning that risk may evolve progressively across turns. Attackers can exploit this by constructing conversations in which individual turns appear benign while the accumulated context gradually steers the model toward harmful compliance. These observations motivate inference-time defenses that explicitly incorporate conversational history and evolving risk signals across turns.

To address this challenge, we propose \textsc{THRD}, a training-free framework for multi-turn jailbreak defense. \textsc{THRD} tracks how instantaneous risk, cumulative intent, and response-level facilitation evolve throughout a conversation, enabling effective inference-time defense against adaptive multi-turn jailbreak attacks. Specifically, \textsc{THRD} decomposes conversational safety assessment into four functional components: a Turn-level Risk Assessor that captures current-turn risk signals, a Historical Context Analyzer that analyzes cumulative intent across turns, a Response Evaluator that assesses whether generated outputs may facilitate downstream harm, and a Decision Module that aggregates these signals through a lightweight temporal scoring mechanism. Rather than relying solely on isolated turn-level judgments, \textsc{THRD} combines current-turn and historical signals within a unified inference-time framework. Extensive experiments demonstrate that \textsc{THRD} achieves strong defense performance while preserving model utility.

\section{Related Work}
\label{app:related}

\subsection{Multi-turn Jailbreak Attacks on LLMs}
Multi-turn jailbreak attacks exploit conversational dynamics to systematically erode safety guardrails across multiple interaction turns. Unlike single-turn attacks that rely on engineered prompts, multi-turn methods leverage conversational coherence to gradually accumulate partial compliance. Crescendo~\cite{russinovich2025great} pioneered incremental escalation from benign questions toward harmful topics. ActorAttack~\cite{ren2024derail} incorporated actor-network theory to discover diverse attack paths while concealing harmful intent. Tempest~\cite{zhou2025tempest} employs breadth-first tree search to explore multiple adversarial paths simultaneously, tracking how minor concessions accumulate into policy violations. RACE~\cite{ying-etal-2025-reasoning} reformulates harmful queries into benign reasoning tasks through attack state machines. X-Teaming~\cite{rahman2025xteaming} introduces a multi-agent framework for collaborative attack planning, optimization, and verification. These methods collectively demonstrate that extended conversations can bypass safety alignment mechanisms designed primarily for single-turn interactions.

\subsection{Defenses Against Jailbreak Attacks}
Existing defenses can be broadly categorized into learning-based and strategy-based approaches. Learning-based methods enhance safety through post-training techniques such as SFT~\cite{NEURIPS2022_b1efde53} or RLHF~\cite{christiano2017deep,korbak2023pretraining,NEURIPS2024_094324f3}. SDGO~\cite{ding-etal-2025-sdgo} leverages the model's discrimination capabilities as reward signals to improve generation safety. MetaDefense~\cite{jiang2025metadefense} introduces a two-stage approach that detects harmful queries before generation and monitors partial responses during generation. Strategy-based methods rely on prompt guidance and content detection without retraining. SAGE~\cite{ding-etal-2025-act} aligns safety discrimination with generation through discriminative analysis and response modules. SafeBehavior~\cite{zhao2025safebehavior} simulates human-like multistage reasoning for jailbreak detection. CCFC~\cite{hu2025ccfc} introduces a dual-track framework that isolates semantic cores to defend against structural attack patterns. While these methods avoid retraining, most focus on single-turn analysis and lack explicit modeling of how risk evolves across conversation turns.

\section{THRD Framework}
We formulate multi-turn jailbreak defense as a \textbf{temporal risk aggregation} problem, where safety depends on how risk accumulates across conversation turns rather than on isolated inputs. This formulation reflects a key insight: in multi-turn interaction, each turn shifts the model's effective conditioning context, and safety behavior becomes trajectory-dependent rather than governed by a fixed decision boundary. As a conversation progresses, earlier turns, including the model's own responses, establish contextual momentum that can progressively lower the barrier to harmful compliance, even when no individual turn is overtly malicious. THRD operationalizes this insight through a modular architecture that prioritizes principled decomposition and task-specific integration over component-level redesign. For analysis mechanisms already validated in prior work, we deliberately adopt and adapt these proven modules rather than re-engineering them for structural novelty alone. The core design effort centers on decomposing the evolving risk trajectory into three complementary signals, adapting each module to its specific role within a temporal defense context, and orchestrating their joint evolution through a time-evolving decision mechanism. As illustrated in Figure~\ref{fig:framework}, THRD decomposes conversational risk into four functional modules corresponding to three complementary dimensions, integrated through a unified scoring framework that captures cross-turn dynamics.

\subsection{Turn-level Risk Assessor}
\label{sec:tra}
The Turn-level Risk Assessor (TRA) estimates \textbf{instantaneous risk} for the current turn, identifying preparatory or escalatory behavior that may signal an emerging attack. Since individual turns may appear benign while contributing to an evolving attack, we adopt the two-stage analysis paradigm from prior work~\cite{ding-etal-2025-act} to capture both semantic and structural signals indicative of emerging threats. Each request is assigned an ordinal score $R_i \in [1, 5]$, serving as a current-state estimate rather than a standalone decision, which provides fine-grained risk gradations to support downstream temporal aggregation.

As the first line of defense in each turn, TRA's scoring feeds into the risk computation and influences how the Decision Module calibrates its response. An overly sensitive TRA propagates false alarms through the pipeline, leading to unnecessary refusals; an insufficiently sensitive TRA allows early-stage attack signals to pass undetected, undermining downstream temporal aggregation. This gatekeeping role makes TRA's calibration particularly critical. We explore three prompt variants beyond the base two-stage analysis to address this calibration challenge: \textbf{Variant A} adds explicit guidance that individual sensitive keywords do not indicate risk and enumerates high-risk structural patterns; \textbf{Variant B} extends Variant A by requiring both benign and harmful interpretations before scoring, defaulting to a lower score unless concrete structural evidence supports the harmful reading; \textbf{Variant C} adds only a single instruction to focus on request structure rather than individual keywords. The impact of these variants is discussed in Section~\ref{sec:main_results}, with details in Appendix.

\subsection{Historical Context Analyzer}
The Historical Context Analyzer (HCA) infers \textbf{cumulative intent} by analyzing the conversation as a temporally ordered interaction. It identifies cross-turn escalation and coordination patterns that are invisible at the single-turn level. To capture these patterns, we analyze conversation structure through three dimensions: \textbf{conversational form characteristics} (e.g., role-play setups), \textbf{evolution patterns} (topic sensitivity progression), and \textbf{attack chain completeness} (whether prior turns form a progression toward harmful objectives), and adopt the multi-stage framework from ARMOR~\cite{zhao2025armoraligningsecuresafe} for semantic reasoning, including intent extraction, strategy identification, and safety judgment. HCA produces a binary judgment $J_h \in \{\text{SAFE}, \text{UNSAFE}\}$ indicating whether the conversation trajectory exhibits malicious intent. Details are provided in Appendix.

\subsection{Response Evaluator}
The Response Evaluator (RE) assesses whether a generated response may \textbf{facilitate downstream harm}, even when it complies with policy. This design is motivated by the discriminator--generator gap~\cite{ding-etal-2025-sdgo}, where models identify harmful content more effectively during evaluation than during generation, making response-level assessment essential for mitigating progressive multi-turn attacks. We evaluate response safety across three dimensions: \textbf{content safety}, \textbf{information sensitivity}, and \textbf{boundary compliance}. RE operates after the current turn's response has been issued and feeds its judgment into the Decision Module at the next turn (Eq.~\ref{eq:risk_score}), rather than intercepting the current response. This post-hoc design reflects the cumulative nature of multi-turn attacks: a single partially facilitative response rarely constitutes a complete safety violation, but it signals that the conversation trajectory is shifting toward harm. By elevating the risk score at the next turn, THRD ensures timely intervention before the attack chain reaches completion, while avoiding the computational cost of generate-then-filter within each turn. RE employs conditional triggering: it activates when $R_i$ indicates medium risk, when $R_i$ is high but $J_h = \text{SAFE}$, or during late conversation stages (details in Appendix. RE outputs a binary judgment $J_r \in \{\text{MAY\_FACILITATE}, \text{WILL\_NOT\_FACILITATE}\}$ for the Decision Module. Details are provided in Appendix.

\begin{table*}[t]
\centering
\resizebox{\textwidth}{!}{
\begin{tabular}{llcccccccc}
\toprule
& & \multicolumn{4}{c}{\textbf{ASR ($\downarrow$)}} & \multicolumn{2}{c}{\textbf{Utility ($\uparrow$)}} & \textbf{Over-Ref. ($\downarrow$)} \\
\cmidrule(lr){3-6} \cmidrule(lr){7-8} \cmidrule(lr){9-9}
& & \multicolumn{2}{c}{X-Teaming} & \multicolumn{2}{c}{Tempest} & & & \\
\cmidrule(lr){3-4} \cmidrule(lr){5-6}
\textbf{Model} & \textbf{Defense} & Harm & Adv & Harm & Adv & MMLU & GSM8K & XSTest \\
\midrule
& No Defense & 100\% & 98.0\% & 84.3\% & 90.0\% & \textbf{82.0\%} & \textbf{92.9\%} & \textbf{9.6\%} \\
& PROACT & 63.5\% {\scriptsize\color{blue}$\downarrow$36.5} & 67.0\% {\scriptsize\color{blue}$\downarrow$31.0} & 2.6\% {\scriptsize\color{blue}$\downarrow$81.7} & 3.1\% {\scriptsize\color{blue}$\downarrow$86.9} & 78.6\% {\scriptsize\color{red}$\downarrow$3.4} & 79.2\% {\scriptsize\color{red}$\downarrow$13.7} & 19.7\% {\scriptsize\color{red}$\uparrow$10.1} \\
Qwen2.5-7B & SAGE & 86.2\% {\scriptsize\color{blue}$\downarrow$13.8} & 83.6\% {\scriptsize\color{blue}$\downarrow$14.4} & 4.9\% {\scriptsize\color{blue}$\downarrow$79.4} & 2.8\% {\scriptsize\color{blue}$\downarrow$87.2} & 68.3\% {\scriptsize\color{red}$\downarrow$13.7} & 68.9\% {\scriptsize\color{red}$\downarrow$24.0} & 61.2\% {\scriptsize\color{red}$\uparrow$51.6} \\
& \textbf{Ours} & \textbf{4.0\%} {\scriptsize\color{blue}$\downarrow$96.0} & \textbf{1.3\%} {\scriptsize\color{blue}$\downarrow$96.7} & \textbf{2.1\%} {\scriptsize\color{blue}$\downarrow$82.2} & \textbf{0.5\%} {\scriptsize\color{blue}$\downarrow$89.5} & \underline{81.7\%} {\scriptsize\color{red}$\downarrow$0.3} & \underline{92.6\%} {\scriptsize\color{red}$\downarrow$0.3} & \underline{12.4\%} {\scriptsize\color{red}$\uparrow$2.8} \\
\midrule
& No Defense & 86.7\% & 92.0\% & 84.7\% & 73.0\% & \textbf{72.4\%} & \textbf{83.4\%} & \textbf{16.0\%} \\
& PROACT & 47.4\% {\scriptsize\color{blue}$\downarrow$39.3} & 24.0\% {\scriptsize\color{blue}$\downarrow$68.0} & 2.4\% {\scriptsize\color{blue}$\downarrow$82.3} & 2.5\% {\scriptsize\color{blue}$\downarrow$70.5} & \underline{71.4\%} {\scriptsize\color{red}$\downarrow$1.0} & \underline{82.8\%} {\scriptsize\color{red}$\downarrow$0.6} & \underline{16.8\%} {\scriptsize\color{red}$\uparrow$0.8} \\
Llama3-8B & SAGE & 17.5\% {\scriptsize\color{blue}$\downarrow$69.2} & 21.2\% {\scriptsize\color{blue}$\downarrow$70.8} & 6.7\% {\scriptsize\color{blue}$\downarrow$78.0} & 3.5\% {\scriptsize\color{blue}$\downarrow$69.5} & 60.6\% {\scriptsize\color{red}$\downarrow$11.8} & 74.2\% {\scriptsize\color{red}$\downarrow$9.2} & 99.6\% {\scriptsize\color{red}$\uparrow$83.6} \\
& \textbf{Ours} & \textbf{1.6\%} {\scriptsize\color{blue}$\downarrow$85.1} & \textbf{0.2\%} {\scriptsize\color{blue}$\downarrow$91.8} & \textbf{0.6\%} {\scriptsize\color{blue}$\downarrow$84.1} & \textbf{1.0\%} {\scriptsize\color{blue}$\downarrow$72.0} & 70.9\% {\scriptsize\color{red}$\downarrow$1.5} & 82.5\% {\scriptsize\color{red}$\downarrow$0.9} & 18.8\% {\scriptsize\color{red}$\uparrow$2.8} \\
\bottomrule
\end{tabular}
}
\caption{Defense performance on multi-turn jailbreak attacks. We evaluate on two target models against X-Teaming and Tempest, measuring ASR on HarmBench (Harm) and AdvBench (Adv), model utility via MMLU (0-shot) and GSM8K, and over-refusal rate via XSTest. {\color{blue}Blue} indicates improvement over No Defense, {\color{red}red} indicates degradation.}
\label{tab:main_results}
\end{table*}

\subsection{Decision Module}
The Decision Module integrates $R_i$, $J_h$, and $J_r$ into a \textbf{time-evolving risk score}, enabling responses ranging from normal generation to refusal. This formulation distinguishes isolated risk from sustained escalation, capturing the iterative nature of multi-turn defense.
At each turn $i$, we convert the binary judgments from HCA and RE into numerical scores:
\begin{equation}
\hat{J}_h^{(i-1)} = \begin{cases}
\gamma, & \text{if } J_h^{(i-1)} = \textsc{Unsafe} \\
0, & \text{if } J_h^{(i-1)} = \textsc{Safe}
\end{cases}
\end{equation}
\begin{equation}
\hat{J}_r^{(i-1)} = \begin{cases}
\gamma, & \text{if } J_r^{(i-1)} = \text{May Facilitate} \\
0, & \text{if } J_r^{(i-1)} = \text{Will Not Facilitate}
\end{cases}
\end{equation}
The risk score for turn $i$ is computed as:
\begin{equation}
F^{(i)} = R_i + \Lambda(R_i) \cdot \Omega(\hat{J}_h^{(i-1)}, \hat{J}_r^{(i-1)})
\label{eq:risk_score}
\end{equation}
where $R_i$ is the current-turn risk score from TRA, and $\Lambda(R_i) = 1 - \frac{R_i - 1}{4}$ is an attenuation function that modulates historical signal influence based on current-turn severity. The composite function $\Omega$ aggregates historical and response-level risks:
\begin{equation}
\Omega(\hat{J}_h^{(i-1)}, \hat{J}_r^{(i-1)}) = \sqrt{\alpha (\hat{J}_h^{(i-1)})^2 + \beta (\hat{J}_r^{(i-1)})^2}
\end{equation}
To capture progressive attack patterns, we apply a trend-based adjustment when risk scores exhibit monotonic increase across consecutive turns ($F^{(i-2)} \leq F^{(i-1)} \leq F^{(i)}$), adding a penalty of $+\delta$ to the risk score. The final adjusted score $F^{(i)}_{\text{adj}}$ is mapped to three risk levels: \emph{low} ($\leq \tau_{low}$), \emph{medium} ($\tau_{low} < F^{(i)}_{\text{adj}} \leq \tau_{high}$), and \emph{high} ($> \tau_{high}$), corresponding to normal generation, controlled generation with safety constraints, and direct refusal, respectively. Once a refusal is triggered at any turn, all subsequent turns are refused unconditionally without further evaluation. This irreversible escalation reflects a key property of multi-turn attacks: once the conversation trajectory has been identified as malicious, subsequent turns---regardless of their surface-level content---are likely continuations of the same attack chain, and re-evaluating them would only provide opportunities for the attacker to craft evasive follow-ups. This design prioritizes security-oriented deployment settings, where preventing recovery from false negatives is considered more critical than recovering from occasional false positives. Detailed response strategies are provided in Appendix, with hyperparameters in Appendix and the complete decision process detailed in Algorithm (Appendix).

\section{Experiments}

\subsection{Experimental Setup}
\label{sec:setup}

\paragraph{Target Models and Infrastructure.}
We evaluate on two target models: Qwen2.5-7B-Instruct~\cite{qwen2025qwen25technicalreport} and Llama-3-8B-Instruct~\cite{dubey2024llama}, served locally via SGLang on a single NVIDIA RTX 4090 GPU. We use Qwen3-32B~\citep{yang2025qwen3} as the attacker model and GPT-4o~\cite{openai2024gpt4technicalreport} as the safety judge.
 
\paragraph{Attacks.}
We test against two state-of-the-art multi-turn attacks: X-Teaming~\cite{rahman2025xteaming}, a multi-agent collaborative framework where GPT-4o~\cite{openai2024gpt4technicalreport} generates attack strategies following the original protocol, and Tempest~\citep{zhou2025tempest}, which employs breadth-first tree search with a branch factor of 5. We also include the single-turn attack AutoDAN~\citep{liu2024autodan} to assess generalization beyond multi-turn settings.
 
\paragraph{Baselines.}
We compare against two inference-time defenses originally designed for single-turn settings: PROACT~\cite{zhao2025proactive} and SAGE~\cite{ding-etal-2025-act}. Since these methods do not explicitly model persistent cross-turn risk, we apply them independently at each conversation turn to simulate multi-turn deployment.

\paragraph{Evaluation.}
Safety is measured by Attack Success Rate (ASR\footnote{Since THRD generates substantive safe responses rather than template-based refusals, we use GPT-4o as the safety judge. Manual annotation on randomly sampled conversations confirms high consistency with GPT-4o's assessments. Keyword-based metrics systematically overestimate ASR in this setting; see Appendix for details.})~\citep{zou2023universal} on HarmBench~\citep{mazeika2024harmbench} and 100 sampled instructions from AdvBench~\citep{zou2023universal}. Utility is assessed via MMLU~\citep{hendrycks2020measuring} (0-shot, 11 selected domains) and GSM8K~\citep{cobbe2021training}. Over-defensiveness is measured by XSTest~\citep{rottger2024xstest}.

\subsection{Main Results}
\label{sec:main_results}

\paragraph{Defense Performance.}
Table~\ref{tab:main_results} presents the overall defense performance. Since no existing defense specifically targets multi-turn jailbreak attacks, we apply two single-turn inference-time defenses---PROACT and SAGE---independently at each conversation turn to simulate multi-turn defense.

A notable pattern emerges across baselines: all defenses perform reasonably well on Tempest, but diverge sharply on X-Teaming. PROACT defends well against Tempest (2.6\% and 3.1\% on Qwen; 2.4\% and 2.5\% on Llama) and maintains low over-refusal (19.7\% on Qwen, 16.8\% on Llama), but remains highly vulnerable to X-Teaming (63.5\% and 67.0\% on Qwen; 47.4\% and 24.0\% on Llama)---indicating that low over-refusal does not compensate for weak attack detection. SAGE achieves moderate ASR on X-Teaming on Llama (17.5\% and 21.2\%) but fails catastrophically on Qwen (86.2\% and 83.6\%), and incurs severe over-refusal (61.2\% on Qwen, 99.6\% on Llama) and substantial utility loss (MMLU drops of 13.7\% and 11.8\%). This divergence suggests that Tempest's tree-search strategy produces attack patterns detectable even by per-turn analysis, while X-Teaming's multi-agent coordination generates more sophisticated cross-turn escalation that exposes the limitations of single-turn defenses applied independently at each turn.

THRD achieves consistently low ASR across both attacks on Qwen (4.0\% and 1.3\% on X-Teaming; 2.1\% and 0.5\% on Tempest) and Llama (1.6\% and 0.2\% on X-Teaming; 0.6\% and 1.0\% on Tempest), while maintaining utility within 1.5\% of the undefended baseline on both models (domain-wise results in Appendix. THRD is the only defense that simultaneously achieves low ASR across both attack types, preserves model utility, and maintains acceptable over-refusal rates, demonstrating that explicit temporal risk modeling can address the trade-off between defense effectiveness and usability that single-turn methods struggle to balance. For over-refusal, analysis of false positives reveals that the primary source is TRA's sensitivity to individual keywords in benign queries (e.g., kill'' in kill the process''), rather than the temporal aggregation mechanism itself. We design three TRA prompt variants (Section~\ref{sec:tra}) that shift scoring emphasis from keyword-level content to request-level structure; the best-performing variant achieves 12.4\% on Qwen, narrowing the gap to only 2.8 percentage points above the undefended baseline (9.6\%). On Llama, over-refusal (18.8\%) leaves room for further calibration. Detailed comparison across all variants is provided in Appendix.

\paragraph{Single-Turn Defense Effectiveness.}
To verify that multi-turn defense does not compromise single-turn robustness, we further evaluate against AutoDAN~\citep{liu2024autodan}, a representative single-turn attack. As shown in Table~\ref{tab:autodan}, THRD, SAGE, and PROACT all achieve 0\% ASR on both models, confirming that the additional temporal aggregation introduced by THRD does not interfere with the target model's ability to handle single-turn adversarial inputs.

\begin{table}[t]
\centering
\small
\begin{tabular}{llcc}
\toprule
\textbf{Model} & \textbf{Defense} & \textbf{Harm} & \textbf{Adv} \\
\midrule
\multirow{4}{*}{Qwen2.5-7B}
& No Defense & 76.9\% & 81.0\% \\
& PROACT & 0 & 0 \\
& SAGE & 0 & 0 \\
& \textbf{Ours} & \textbf{0} & \textbf{0} \\
\midrule
\multirow{4}{*}{Llama3-8B}
& No Defense & 60.0\% & 75.0\% \\
& PROACT & 0 & 0 \\
& SAGE & 0 & 0 \\
& \textbf{Ours} & \textbf{0} & \textbf{0} \\
\bottomrule
\end{tabular}
\caption{ASR ($\downarrow$) against AutoDAN (single-turn attack). All three defenses achieve complete protection, confirming that multi-turn defense mechanisms preserve single-turn robustness.}
\label{tab:autodan}
\end{table}

\begin{table}[t]
\centering
\small
\begin{tabular}{lc}
\toprule
\textbf{Configuration} & \textbf{ASR} $\downarrow$ \\
\midrule
Full Framework (THRD) & 1.59\% \\
\midrule
\quad w/o Persistent Rejection & 5.15\% {\scriptsize\color{red}$\uparrow$3.56} \\
\quad w/o TRA & 25.37\% {\scriptsize\color{red}$\uparrow$23.78}
\\
\quad w/o HCA & 26.07\% {\scriptsize\color{red}$\uparrow$24.48} \\
\quad w/o RE & 17.07\% {\scriptsize\color{red}$\uparrow$15.48} \\
\quad w/o HCA \& RE & 33.30\% {\scriptsize\color{red}$\uparrow$31.71} \\
\bottomrule
\end{tabular}
\caption{Ablation study on X-Teaming (Llama3-8B, HarmBench). {\color{red}Red}: degradation compared to Full Framework. Persistent Rejection operates only after a high-risk refusal is triggered and does not affect single-turn over-refusal measurements.}
\label{tab:ablation}
\end{table}

\subsection{Ablation Study}
\label{sec:ablation}
 
\paragraph{Module Ablation.}

Table~\ref{tab:ablation} presents the contribution of each component on X-Teaming (Llama3-8B, HarmBench). Removing Persistent Rejection increases ASR from 1.59\% to 5.15\%, confirming that attackers actively attempt to recover from rejected turns through evasive follow-ups. Among the three analysis modules, removing TRA or HCA causes the largest and comparable degradation (+23.78\% and +24.48\%), confirming that current-turn and cross-turn analysis are equally critical and non-redundant. Removing RE yields a smaller but substantial degradation (+15.48\%), demonstrating its complementary role in detecting response-level facilitation that sustains attack trajectories even when TRA and HCA correctly identify risk. Removing both HCA and RE produces the worst performance (+31.71\%), though TRA alone still reduces ASR from the undefended 86.7\% to 33.30\%, indicating meaningful baseline protection from turn-level assessment.

\begin{table}[t]
\centering
\small
\begin{tabular}{lcc}
\toprule
\textbf{Module} & \textbf{Qwen2.5-7B} & \textbf{Llama3-8B} \\
\midrule
TRA & 3.68s & 3.26s \\
HCA & 9.60s & 17.24s \\
RE & 1.92s & 1.72s \\
Decision & $\approx$0s & $\approx$0s \\
\midrule
THRD Total & $\approx$15.2s & $\approx$22.2s \\
\midrule
SAGE & 5.37s & 8.27s \\
\bottomrule
\end{tabular}
\caption{Per-turn single-call latency (mean).}
\label{tab:overhead}
\end{table}

\subsection{Analysis}
\label{sec:analysis}

\paragraph{Computational Overhead.}
Table~\ref{tab:overhead} reports per-module single-call latency measured on a single RTX 4090 GPU. HCA is the dominant bottleneck, accounting for 63.2\% of total overhead on Qwen2.5-7B (9.60s / 15.2s) and 77.7\% on Llama3-8B (17.24s / 22.2s), as it performs cross-turn semantic analysis over the full conversation history. The higher HCA latency on Llama3-8B likely results from longer unblocked responses expanding the context. Compared to SAGE (5.37s on Qwen, 8.27s on Llama), THRD's total latency of 15--22s is higher, but this trades controlled latency for substantially stronger defense: SAGE fails catastrophically on X-Teaming (Table~\ref{tab:main_results}) despite lower overhead. Since HCA dominates the overall latency, future optimization efforts could focus on reducing its input length through context compression techniques, or merging TRA and HCA into a unified module that jointly performs current-turn and historical analysis in a single inference call.

\begin{table}[t]
\centering
\small
\begin{tabular}{lcc}
\toprule
\textbf{History Order} & \textbf{Qwen2.5-7B} & \textbf{Llama3-8B} \\
\midrule
Ordered & 0.7\% & 0.8\% \\
Shuffled & 1.3\% & 1.2\% \\
\bottomrule
\end{tabular}
\caption{Effect of shuffling conversation history on ASR (Tempest, AdvBench). Disrupting temporal order degrades defense on both models.}
\label{tab:shuffle}
\end{table}

\begin{table}[t]
\centering
\small
\begin{tabular}{llcc}
\toprule
$\Lambda$ & $\Omega$ & \textbf{Qwen} & \textbf{Llama} \\
\midrule
Linear Decay & L2 Norm & 0.57\% & \textbf{0.85\%} \\
\midrule
\multicolumn{4}{l}{\textit{$\Lambda$ variants}} \\
No Decay & L2 Norm & 0.32\% & 1.91\% \\
Half Decay & L2 Norm & 1.14\% & 1.96\% \\
Inverse Decay & L2 Norm & 0.69\% & 1.93\% \\
\midrule
\multicolumn{4}{l}{\textit{$\Omega$ variants}} \\
Linear Decay & Weighted Sum & \textbf{0.25\%} & 1.53\% \\
Linear Decay & Max & 0.83\% & 1.15\% \\
Linear Decay & Geo.\ Mean & 0.47\% & 1.79\% \\
\bottomrule
\end{tabular}
\caption{Decision formula ablation (ASR $\downarrow$). The first row is the shared configuration across both ablation groups. Evaluated on Tempest with 100 AdvBench instructions.}
\label{tab:formula_ablation}
\end{table}

\paragraph{History Order Sensitivity.}
To verify that HCA relies on temporal ordering rather than bag-of-turns content, we shuffle the conversation history before passing it to HCA while keeping all other modules unchanged. Table~\ref{tab:shuffle} shows the results on Tempest (AdvBench). Shuffling increases ASR on both models (Qwen: 0.7\%$\rightarrow$1.3\%; Llama: 0.8\%$\rightarrow$1.2\%), confirming that HCA exploits sequential structure---the ordering of turns establishes contextual momentum that progressively conditions the model toward compliance, and disrupting this ordering weakens the accumulated conditioning effect. However, the shuffled configuration still maintains low absolute ASR ($\leq$1.3\%), indicating that conversation history also contains valuable content-level signals---such as recurring sensitive topics and cumulative intent---that HCA can leverage even without temporal ordering. This validates our design choice of retaining the full conversation history: temporal structure provides critical ordering cues, while the content of earlier turns contributes complementary evidence, and the combination of both enables THRD's robust cross-turn defense.

\paragraph{Decision Formula Analysis.}
We ablate the attenuation function $\Lambda$ and aggregation function $\Omega$ in the Decision Module. We denote the four $\Lambda$ variants as: \textit{Linear Decay} ($1 - \frac{R_i - 1}{R_{\max} - 1}$), \textit{No Decay} ($\Lambda = 1$), \textit{Half Decay} ($\Lambda = 0.5$), and \textit{Inverse Decay} ($\frac{R_i - 1}{R_{\max} - 1}$). The four $\Omega$ variants are: \textit{L2 Norm} ($\sqrt{\alpha \hat{J}_h^2 + \beta \hat{J}_r^2}$), \textit{Weighted Sum} ($\alpha \hat{J}_h + \beta \hat{J}_r$), \textit{Max} ($\max(\alpha \hat{J}_h, \beta \hat{J}_r)$), and \textit{Geometric Mean} ($\sqrt{\hat{J}_h \cdot \hat{J}_r}$). All experiments are conducted on Tempest with 100 AdvBench instructions.

Table~\ref{tab:formula_ablation} shows the results. All seven configurations achieve low absolute ASR ($<$2\%), confirming robustness to specific functional choices. However, the variants differ in \textbf{cross-architecture consistency}: most alternatives achieve low ASR on Qwen but degrade on Llama (e.g., No Decay: 0.32\% vs.\ 1.91\%; Weighted Sum: 0.25\% vs.\ 1.53\%). Linear Decay paired with L2 Norm yields the most stable performance across both models (0.57\% and 0.85\%), with the smallest cross-architecture gap of 0.28 percentage points.

\begin{figure}[t]
\centering
\includegraphics[width=\columnwidth]{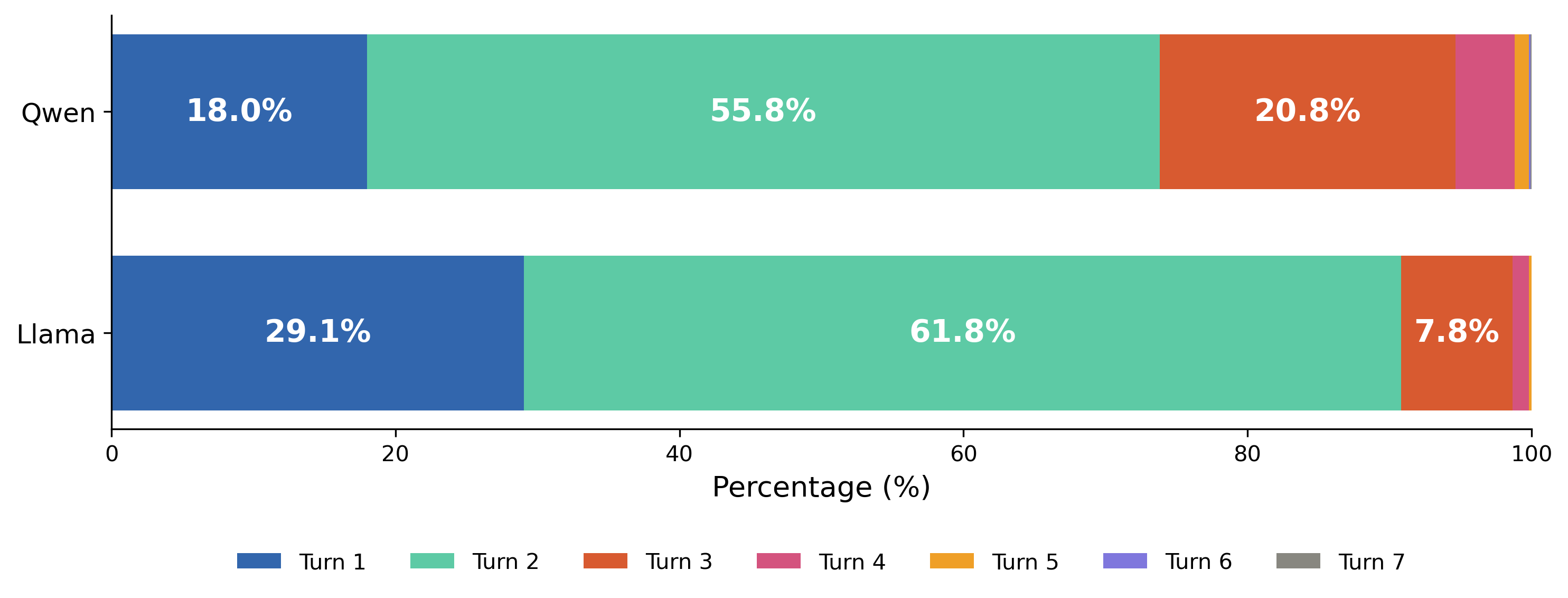}
\caption{Distribution of first defense trigger turn on X-Teaming (up to 7 turns). Over 70\% of attacks require Turn~2 or later to detect.}
\label{fig:first_trigger_xteaming}
\end{figure}

\paragraph{First Rejection Trigger Analysis.}
A natural question is whether single-turn detection suffices or multi-turn awareness is genuinely necessary. Figure~\ref{fig:first_trigger_xteaming} shows the distribution of the turn at which THRD first triggers a rejection on X-Teaming (up to 7 turns). Only 18.0\% (Qwen) and 29.1\% (Llama) of attacks are caught at Turn~1; the majority are first detected at Turn~2 (55.8\% and 61.8\%), with a non-trivial portion requiring Turn~3 or later (26.2\% and 9.1\%). This concentration at Turn~2 highlights that multi-turn attacks deliberately design the first turn to appear benign, deferring boundary-probing to subsequent turns---explaining why over 70\% of attacks evade first-turn detection. However, nearly all attacks are detected within the first three turns, suggesting that current multi-turn attack methods exhibit relatively predictable escalation patterns. More sophisticated attacks that distribute intent more gradually across longer trajectories would pose a greater challenge, motivating continued research on both fronts.

\begin{figure}[t]
    \centering
    \includegraphics[width=\columnwidth]{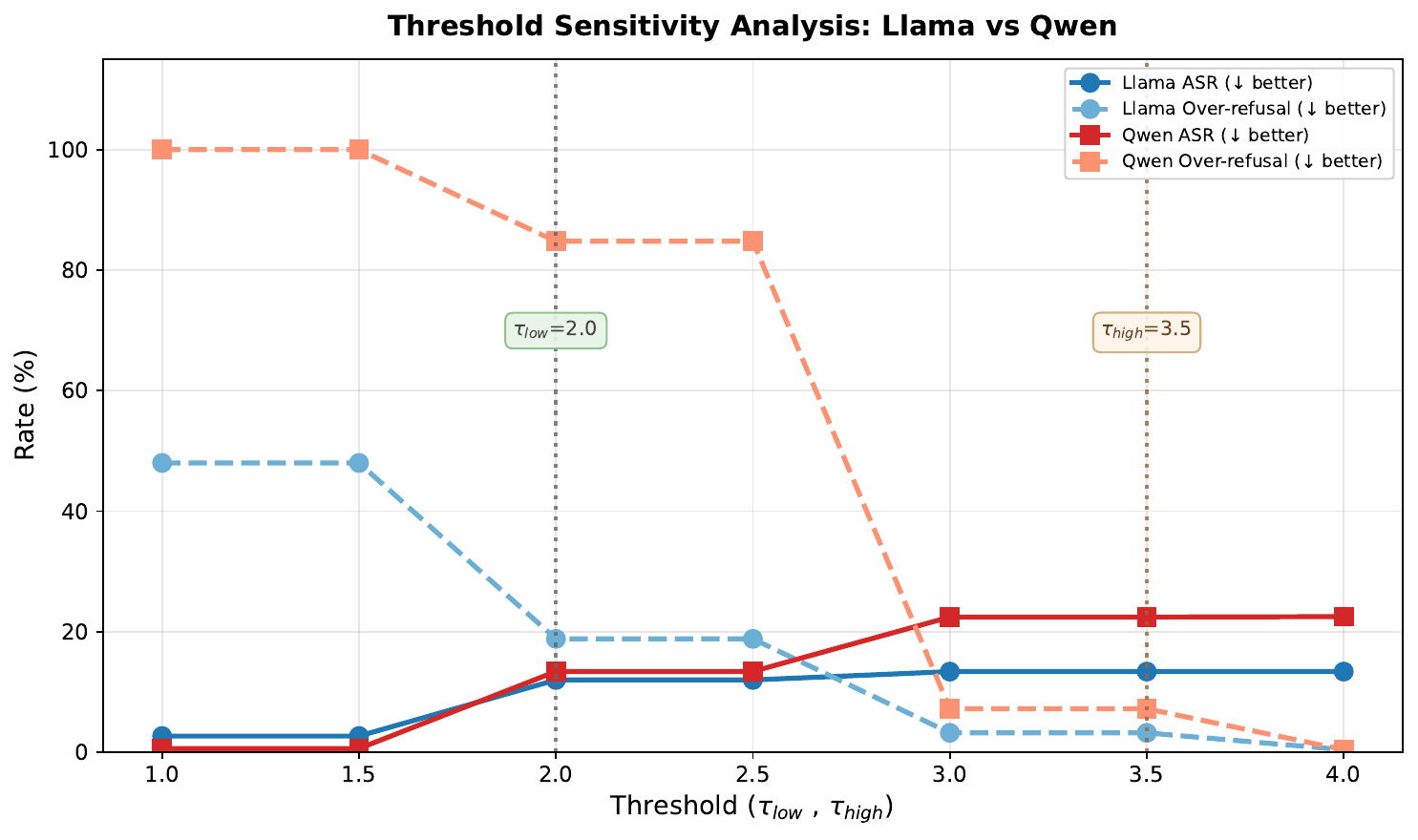}
    \caption{Threshold sensitivity analysis. ASR ($\downarrow$ better) and over-refusal rate ($\downarrow$ better) under different threshold values. $\tau_{\text{low}}=2.0$ and $\tau_{\text{high}}=3.5$ are selected as optimal trade-off points.}
    \label{fig:threshold}
\end{figure}

\paragraph{Threshold and Risk Dynamics.}
Figure~\ref{fig:threshold} shows ASR and over-refusal rate as a function of the decision threshold. ASR is defined as the proportion of attack conversations whose final-turn risk score falls below the threshold, while over-refusal measures the proportion of benign queries incorrectly flagged above it. Attack conversations consistently receive high risk scores: even at the lowest threshold ($\tau=1.0$), ASR remains below 3\% on both models, indicating that THRD assigns elevated scores to nearly all attack trajectories. Over-refusal, however, is highly sensitive to the threshold---dropping sharply from near 100\% at $\tau=1.0$ to below 20\% at $\tau=2.0$ on both models. This confirms that the primary challenge lies not in detecting attacks but in avoiding false positives on benign inputs, and motivates our selection of $\tau_{\text{low}}=2.0$ and $\tau_{\text{high}}=3.5$ as the operating points that balance these two objectives.

\begin{figure}[t]
\centering
\includegraphics[width=\columnwidth]{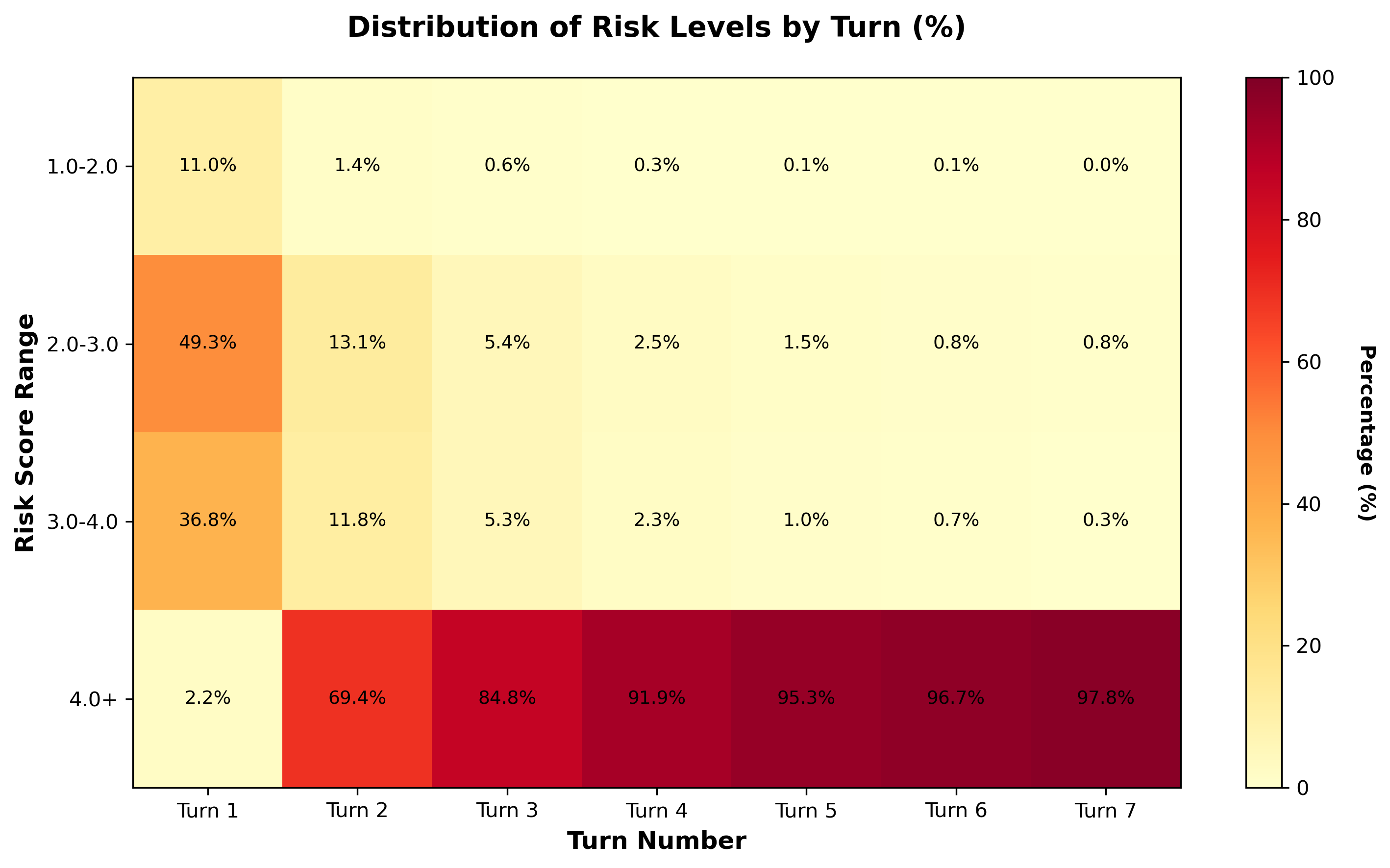}
\caption{Distribution of risk levels across conversation turns. High-risk conversations increase significantly after Turn 1.}
\label{fig:risk_heatmap}
\end{figure}

Figure~\ref{fig:risk_heatmap} further illustrates how risk scores evolve across conversation turns. At Turn~1, scores are distributed across all levels---49.3\% in the 2.0--3.0 range and 36.8\% in 3.0--4.0---reflecting the ambiguity of initial turns designed to appear benign. From Turn~2 onward, scores rapidly concentrate in the 4.0+ range (69.4\%$\rightarrow$84.8\%$\rightarrow$91.9\%), demonstrating that THRD's temporal scoring mechanism effectively captures attack escalation once cross-turn patterns emerge. Additional visualizations are provided in Appendix.

\section{Conclusion}
\label{sec:conclusion}

We present \textsc{THRD}, a training-free framework for defending against multi-turn jailbreak attacks. \textsc{THRD} combines turn-level risk assessment, historical context analysis, response evaluation, and lightweight cross-turn risk aggregation within a unified inference-time framework. By incorporating both current-turn and historical signals, \textsc{THRD} effectively detects progressive attack patterns that are inherently difficult to identify through isolated single-turn analysis. Experiments against state-of-the-art multi-turn attacks---including tree-search-based and multi-agent collaborative methods---across multiple target models show that \textsc{THRD} achieves consistently low attack success rates while largely preserving model utility. Ablation studies further confirm that each module provides non-redundant contributions, and that the decision formula generalizes stably across model architectures. Further analysis reveals that over 70\% of multi-turn attacks evade first-turn detection, underscoring the practical necessity of incorporating cross-turn temporal information in jailbreak defense. Future work may explore context compression and module unification to reduce computational overhead, as well as adaptive or lightweight learned components to further improve the balance between defense effectiveness and over-refusal.

\section*{Limitations}

THRD adopts a modular inference-time design that explicitly models conversational risk accumulation without additional training. While empirically robust against current attacks, several limitations remain. The TRA, HCA, and RE modules rely on prompt-based semantic judgments, making the framework partially dependent on the underlying evaluator model's reasoning ability. Future attacks specifically optimized against THRD's prompting patterns or scoring dynamics may reduce effectiveness. Relatedly, the current implementation models temporal dynamics through lightweight score aggregation rather than sequential learning, which may limit effectiveness against attacks distributing intent across substantially longer trajectories. The persistent rejection mechanism intentionally prioritizes security-critical settings but may be overly conservative for general-purpose systems; reversible or uncertainty-aware recovery mechanisms could better balance security and interaction continuity. Finally, multi-module analysis over conversation history introduces inference overhead, with HCA as the primary bottleneck. Context compression, incremental history updating, or module unification may help address this in future work.

\section*{Ethical Considerations}
Our research is committed to enhancing the safety of LLMs by addressing vulnerabilities to multi-turn jailbreak attacks through a training-free defense framework. We emphasize that our work is grounded in ethical considerations, aiming to mitigate the generation of harmful content rather than introducing new risks. All jailbreak attack methods and datasets used in our experiments are derived from publicly available sources (X-Teaming, Tempest, AutoDAN, HarmBench, and AdvBench), ensuring transparency and avoiding the introduction of new attack vectors. Our findings demonstrate that our proposed framework significantly reduces attack success rates across different models and attack strategies, thereby promoting the responsible deployment of LLMs. We acknowledge that the development of any defense mechanism may inspire adversarial adaptations; however, our primary focus remains on safeguarding LLMs from existing and emerging multi-turn threats. We believe that our work contributes to the development of more robust and trustworthy AI systems. By sharing our methodologies and findings, we aim to support the broader research community in building safer LLM applications.

\bibliography{custom}

@article{christiano2017deep,
  title={Deep reinforcement learning from human preferences},
  author={Christiano, Paul F and Leike, Jan and Brown, Tom and Martic, Miljan and Legg, Shane and Amodei, Dario},
  journal={Advances in neural information processing systems},
  volume={30},
  year={2017}
}

@inproceedings{russinovich2025great,
  title={Great, Now Write an Article About That: The Crescendo Multi-Turn LLM Jailbreak Attack},
  author={Russinovich, Mark and Salem, Ahmed and Eldan, Ronen},
  booktitle={34th USENIX Security Symposium (USENIX Security 25)},
  pages={2421--2440},
  year={2025}
}

@article{ren2024derail,
  title={Derail yourself: Multi-turn llm jailbreak attack through self-discovered clues},
  author={Ren, Qibing and Li, Hao and Liu, Dongrui and Xie, Zhanxu and Lu, Xiaoya and Qiao, Yu and Sha, Lei and Yan, Junchi and Ma, Lizhuang and Shao, Jing},
  year={2024}
}

@inproceedings{
zhou2025tempest,
title={{TEMPEST}: Multi-Turn Jailbreaking of Large Language Models with Tree Search},
author={Andy Zhou and Ron Arel},
booktitle={ICLR 2025 Workshop on Building Trust in Language Models and Applications},
year={2025},
url={https://openreview.net/forum?id=rDC2UVdB0t}
}

@inproceedings{ying-etal-2025-reasoning,
    title = "Reasoning-Augmented Conversation for Multi-Turn Jailbreak Attacks on Large Language Models",
    author = "Ying, Zonghao  and
      Zhang, Deyue  and
      Jing, Zonglei  and
      Xiao, Yisong  and
      Zou, Quanchen  and
      Liu, Aishan  and
      Liang, Siyuan  and
      Zhang, Xiangzheng  and
      Liu, Xianglong  and
      Tao, Dacheng",
    editor = "Christodoulopoulos, Christos  and
      Chakraborty, Tanmoy  and
      Rose, Carolyn  and
      Peng, Violet",
    booktitle = "Findings of the Association for Computational Linguistics: EMNLP 2025",
    month = nov,
    year = "2025",
    address = "Suzhou, China",
    publisher = "Association for Computational Linguistics",
    url = "https://aclanthology.org/2025.findings-emnlp.929/",
    doi = "10.18653/v1/2025.findings-emnlp.929",
    pages = "17138--17157",
    ISBN = "979-8-89176-335-7"
}

@inproceedings{
rahman2025xteaming,
title={X-Teaming: Multi-Turn Jailbreaks and Defenses with Adaptive Multi-Agents},
author={Salman Rahman and Liwei Jiang and James Shiffer and Genglin Liu and Sheriff Issaka and Md Rizwan Parvez and Hamid Palangi and Kai-Wei Chang and Yejin Choi and Saadia Gabriel},
booktitle={Second Conference on Language Modeling},
year={2025},
url={https://openreview.net/forum?id=gKfj7Jb1kj}
}

@article{dubey2024llama,
  title={The llama 3 herd of models},
  author={Dubey, Abhimanyu and Jauhri, Abhinav and Pandey, Abhinav and Kadian, Abhishek and Al-Dahle, Ahmad and Letman, Aiesha and Mathur, Akhil and Schelten, Alan and Yang, Amy and Fan, Angela and others},
  journal={arXiv e-prints},
  pages={arXiv--2407},
  year={2024}
}

@inproceedings{korbak2023pretraining,
  title={Pretraining language models with human preferences},
  author={Korbak, Tomasz and Shi, Kejian and Chen, Angelica and Bhalerao, Rasika Vinayak and Buckley, Christopher and Phang, Jason and Bowman, Samuel R and Perez, Ethan},
  booktitle={International Conference on Machine Learning},
  pages={17506--17533},
  year={2023},
  organization={PMLR}
}

@inproceedings{ding-etal-2025-sdgo,
    title = "{SDGO}: Self-Discrimination-Guided Optimization for Consistent Safety in Large Language Models",
    author = "Ding, Peng  and
      Sun, Wen  and
      Li, Dailin  and
      Zou, Wei  and
      Wang, Jiaming  and
      Chen, Jiajun  and
      Huang, Shujian",
    editor = "Christodoulopoulos, Christos  and
      Chakraborty, Tanmoy  and
      Rose, Carolyn  and
      Peng, Violet",
    booktitle = "Proceedings of the 2025 Conference on Empirical Methods in Natural Language Processing",
    month = nov,
    year = "2025",
    address = "Suzhou, China",
    publisher = "Association for Computational Linguistics",
    url = "https://aclanthology.org/2025.emnlp-main.253/",
    doi = "10.18653/v1/2025.emnlp-main.253",
    pages = "5023--5037",
    ISBN = "979-8-89176-332-6"
}

@inproceedings{
jiang2025metadefense,
title={MetaDefense: Defending Fine-tuning based Jailbreak Attack Before and During Generation},
author={Weisen Jiang and Sinno Jialin Pan},
booktitle={The Thirty-ninth Annual Conference on Neural Information Processing Systems},
year={2025},
url={https://openreview.net/forum?id=ycMpNwzUAA}
}

@inproceedings{ding-etal-2025-act,
    title = "Why Not Act on What You Know? Unleashing Safety Potential of {LLM}s via Self-Aware Guard Enhancement",
    author = "Ding, Peng  and
      Kuang, Jun  and
      Wang, ZongYu  and
      Cao, Xuezhi  and
      Cai, Xunliang  and
      Chen, Jiajun  and
      Huang, Shujian",
    editor = "Che, Wanxiang  and
      Nabende, Joyce  and
      Shutova, Ekaterina  and
      Pilehvar, Mohammad Taher",
    booktitle = "Findings of the Association for Computational Linguistics: ACL 2025",
    month = jul,
    year = "2025",
    address = "Vienna, Austria",
    publisher = "Association for Computational Linguistics",
    url = "https://aclanthology.org/2025.findings-acl.325/",
    doi = "10.18653/v1/2025.findings-acl.325",
    pages = "6279--6299",
    ISBN = "979-8-89176-256-5"
}

@article{zhao2025safebehavior,
  title={SafeBehavior: Simulating Human-Like Multistage Reasoning to Mitigate Jailbreak Attacks in Large Language Models},
  author={Zhao, Qinjian and Wang, Jiaqi and Gao, Zhiqiang and Dou, Zhihao and Abuhaija, Belal and Huang, Kaizhu},
  journal={arXiv preprint arXiv:2509.26345},
  year={2025}
}

@article{hu2025ccfc,
  title={CCFC: Core \& Core-Full-Core Dual-Track Defense for LLM Jailbreak Protection},
  author={Hu, Jiaming and Wang, Haoyu and Mukherjee, Debarghya and Paschalidis, Ioannis Ch},
  journal={arXiv preprint arXiv:2508.14128},
  year={2025}
}

@misc{qwen2025qwen25technicalreport,
      title={Qwen2.5 Technical Report}, 
      author={Qwen and : and An Yang and Baosong Yang and Beichen Zhang and Binyuan Hui and Bo Zheng and Bowen Yu and Chengyuan Li and Dayiheng Liu and Fei Huang and Haoran Wei and Huan Lin and Jian Yang and Jianhong Tu and Jianwei Zhang and Jianxin Yang and Jiaxi Yang and Jingren Zhou and Junyang Lin and Kai Dang and Keming Lu and Keqin Bao and Kexin Yang and Le Yu and Mei Li and Mingfeng Xue and Pei Zhang and Qin Zhu and Rui Men and Runji Lin and Tianhao Li and Tianyi Tang and Tingyu Xia and Xingzhang Ren and Xuancheng Ren and Yang Fan and Yang Su and Yichang Zhang and Yu Wan and Yuqiong Liu and Zeyu Cui and Zhenru Zhang and Zihan Qiu},
      year={2025},
      eprint={2412.15115},
      archivePrefix={arXiv},
      primaryClass={cs.CL},
      url={https://arxiv.org/abs/2412.15115}, 
}

@article{yang2025qwen3,
  title={Qwen3 technical report},
  author={Yang, An and Li, Anfeng and Yang, Baosong and Zhang, Beichen and Hui, Binyuan and Zheng, Bo and Yu, Bowen and Gao, Chang and Huang, Chengen and Lv, Chenxu and others},
  journal={arXiv preprint arXiv:2505.09388},
  year={2025}
}

@misc{openai2024gpt4technicalreport,
      title={GPT-4 Technical Report}, 
      author={OpenAI and Josh Achiam and Steven Adler and Sandhini Agarwal and Lama Ahmad and Ilge Akkaya and Florencia Leoni Aleman and Diogo Almeida and Janko Altenschmidt and Sam Altman and Shyamal Anadkat and Red Avila and Igor Babuschkin and Suchir Balaji and Valerie Balcom and Paul Baltescu and Haiming Bao and Mohammad Bavarian and Jeff Belgum and Irwan Bello and Jake Berdine and Gabriel Bernadett-Shapiro and Christopher Berner and Lenny Bogdonoff and Oleg Boiko and Madelaine Boyd and Anna-Luisa Brakman and Greg Brockman and Tim Brooks and Miles Brundage and Kevin Button and Trevor Cai and Rosie Campbell and Andrew Cann and Brittany Carey and Chelsea Carlson and Rory Carmichael and Brooke Chan and Che Chang and Fotis Chantzis and Derek Chen and Sully Chen and Ruby Chen and Jason Chen and Mark Chen and Ben Chess and Chester Cho and Casey Chu and Hyung Won Chung and Dave Cummings and Jeremiah Currier and Yunxing Dai and Cory Decareaux and Thomas Degry and Noah Deutsch and Damien Deville and Arka Dhar and David Dohan and Steve Dowling and Sheila Dunning and Adrien Ecoffet and Atty Eleti and Tyna Eloundou and David Farhi and Liam Fedus and Niko Felix and Simón Posada Fishman and Juston Forte and Isabella Fulford and Leo Gao and Elie Georges and Christian Gibson and Vik Goel and Tarun Gogineni and Gabriel Goh and Rapha Gontijo-Lopes and Jonathan Gordon and Morgan Grafstein and Scott Gray and Ryan Greene and Joshua Gross and Shixiang Shane Gu and Yufei Guo and Chris Hallacy and Jesse Han and Jeff Harris and Yuchen He and Mike Heaton and Johannes Heidecke and Chris Hesse and Alan Hickey and Wade Hickey and Peter Hoeschele and Brandon Houghton and Kenny Hsu and Shengli Hu and Xin Hu and Joost Huizinga and Shantanu Jain and Shawn Jain and Joanne Jang and Angela Jiang and Roger Jiang and Haozhun Jin and Denny Jin and Shino Jomoto and Billie Jonn and Heewoo Jun and Tomer Kaftan and Łukasz Kaiser and Ali Kamali and Ingmar Kanitscheider and Nitish Shirish Keskar and Tabarak Khan and Logan Kilpatrick and Jong Wook Kim and Christina Kim and Yongjik Kim and Jan Hendrik Kirchner and Jamie Kiros and Matt Knight and Daniel Kokotajlo and Łukasz Kondraciuk and Andrew Kondrich and Aris Konstantinidis and Kyle Kosic and Gretchen Krueger and Vishal Kuo and Michael Lampe and Ikai Lan and Teddy Lee and Jan Leike and Jade Leung and Daniel Levy and Chak Ming Li and Rachel Lim and Molly Lin and Stephanie Lin and Mateusz Litwin and Theresa Lopez and Ryan Lowe and Patricia Lue and Anna Makanju and Kim Malfacini and Sam Manning and Todor Markov and Yaniv Markovski and Bianca Martin and Katie Mayer and Andrew Mayne and Bob McGrew and Scott Mayer McKinney and Christine McLeavey and Paul McMillan and Jake McNeil and David Medina and Aalok Mehta and Jacob Menick and Luke Metz and Andrey Mishchenko and Pamela Mishkin and Vinnie Monaco and Evan Morikawa and Daniel Mossing and Tong Mu and Mira Murati and Oleg Murk and David Mély and Ashvin Nair and Reiichiro Nakano and Rajeev Nayak and Arvind Neelakantan and Richard Ngo and Hyeonwoo Noh and Long Ouyang and Cullen O'Keefe and Jakub Pachocki and Alex Paino and Joe Palermo and Ashley Pantuliano and Giambattista Parascandolo and Joel Parish and Emy Parparita and Alex Passos and Mikhail Pavlov and Andrew Peng and Adam Perelman and Filipe de Avila Belbute Peres and Michael Petrov and Henrique Ponde de Oliveira Pinto and Michael and Pokorny and Michelle Pokrass and Vitchyr H. Pong and Tolly Powell and Alethea Power and Boris Power and Elizabeth Proehl and Raul Puri and Alec Radford and Jack Rae and Aditya Ramesh and Cameron Raymond and Francis Real and Kendra Rimbach and Carl Ross and Bob Rotsted and Henri Roussez and Nick Ryder and Mario Saltarelli and Ted Sanders and Shibani Santurkar and Girish Sastry and Heather Schmidt and David Schnurr and John Schulman and Daniel Selsam and Kyla Sheppard and Toki Sherbakov and Jessica Shieh and Sarah Shoker and Pranav Shyam and Szymon Sidor and Eric Sigler and Maddie Simens and Jordan Sitkin and Katarina Slama and Ian Sohl and Benjamin Sokolowsky and Yang Song and Natalie Staudacher and Felipe Petroski Such and Natalie Summers and Ilya Sutskever and Jie Tang and Nikolas Tezak and Madeleine B. Thompson and Phil Tillet and Amin Tootoonchian and Elizabeth Tseng and Preston Tuggle and Nick Turley and Jerry Tworek and Juan Felipe Cerón Uribe and Andrea Vallone and Arun Vijayvergiya and Chelsea Voss and Carroll Wainwright and Justin Jay Wang and Alvin Wang and Ben Wang and Jonathan Ward and Jason Wei and CJ Weinmann and Akila Welihinda and Peter Welinder and Jiayi Weng and Lilian Weng and Matt Wiethoff and Dave Willner and Clemens Winter and Samuel Wolrich and Hannah Wong and Lauren Workman and Sherwin Wu and Jeff Wu and Michael Wu and Kai Xiao and Tao Xu and Sarah Yoo and Kevin Yu and Qiming Yuan and Wojciech Zaremba and Rowan Zellers and Chong Zhang and Marvin Zhang and Shengjia Zhao and Tianhao Zheng and Juntang Zhuang and William Zhuk and Barret Zoph},
      year={2024},
      eprint={2303.08774},
      archivePrefix={arXiv},
      primaryClass={cs.CL},
      url={https://arxiv.org/abs/2303.08774}, 
}

@inproceedings{
mazeika2024harmbench,
title={HarmBench: A Standardized Evaluation Framework for Automated Red Teaming and Robust Refusal},
author={Mantas Mazeika and Long Phan and Xuwang Yin and Andy Zou and Zifan Wang and Norman Mu and Elham Sakhaee and Nathaniel Li and Steven Basart and Bo Li and David Forsyth and Dan Hendrycks},
booktitle={Forty-first International Conference on Machine Learning},
year={2024},
url={https://openreview.net/forum?id=f3TUipYU3U}
}

@article{zou2023universal,
  title={Universal and transferable adversarial attacks on aligned language models},
  author={Zou, Andy and Wang, Zifan and Carlini, Nicholas and Nasr, Milad and Kolter, J Zico and Fredrikson, Matt},
  journal={arXiv preprint arXiv:2307.15043},
  year={2023}
}

@article{hendrycks2020measuring,
  title={Measuring massive multitask language understanding},
  author={Hendrycks, Dan and Burns, Collin and Basart, Steven and Zou, Andy and Mazeika, Mantas and Song, Dawn and Steinhardt, Jacob},
  journal={arXiv preprint arXiv:2009.03300},
  year={2020}
}

@inproceedings{rottger2024xstest,
  title={Xstest: A test suite for identifying exaggerated safety behaviours in large language models},
  author={R{\"o}ttger, Paul and Kirk, Hannah and Vidgen, Bertie and Attanasio, Giuseppe and Bianchi, Federico and Hovy, Dirk},
  booktitle={Proceedings of the 2024 Conference of the North American Chapter of the Association for Computational Linguistics: Human Language Technologies (Volume 1: Long Papers)},
  pages={5377--5400},
  year={2024}
}

@inproceedings{NEURIPS2022_b1efde53,
 author = {Ouyang, Long and Wu, Jeffrey and Jiang, Xu and Almeida, Diogo and Wainwright, Carroll and Mishkin, Pamela and Zhang, Chong and Agarwal, Sandhini and Slama, Katarina and Ray, Alex and Schulman, John and Hilton, Jacob and Kelton, Fraser and Miller, Luke and Simens, Maddie and Askell, Amanda and Welinder, Peter and Christiano, Paul F and Leike, Jan and Lowe, Ryan},
 booktitle = {Advances in Neural Information Processing Systems},
 editor = {S. Koyejo and S. Mohamed and A. Agarwal and D. Belgrave and K. Cho and A. Oh},
 pages = {27730--27744},
 publisher = {Curran Associates, Inc.},
 title = {Training language models to follow instructions with human feedback},
 url = {https://proceedings.neurips.cc/paper_files/paper/2022/file/b1efde53be364a73914f58805a001731-Paper-Conference.pdf},
 volume = {35},
 year = {2022}
}

@misc{openai2025gpt5,
  title={Introducing GPT-5},
  author={OpenAI},
  year={2025},
  url={https://openai.com/index/introducing-gpt-5/},
  note={Accessed: December 2025}
}

@misc{anthropic2025claude4,
  title={Introducing Claude 4},
  author={Anthropic},
  year={2025},
  url={https://www.anthropic.com/news/claude-4},
  note={Accessed: December 2025}
}

@article{ouyang2022training,
  title={Training language models to follow instructions with human feedback},
  author={Ouyang, Long and Wu, Jeffrey and Jiang, Xu and Almeida, Diogo and Wainwright, Carroll and Mishkin, Pamela and Zhang, Chong and Agarwal, Sandhini and Slama, Katarina and Ray, Alex and others},
  journal={Advances in neural information processing systems},
  volume={35},
  pages={27730--27744},
  year={2022}
}

@inproceedings{lin-etal-2024-mitigating,
    title = "Mitigating the Alignment Tax of {RLHF}",
    author = "Lin, Yong  and
      Lin, Hangyu  and
      Xiong, Wei  and
      Diao, Shizhe  and
      Liu, Jianmeng  and
      Zhang, Jipeng  and
      Pan, Rui  and
      Wang, Haoxiang  and
      Hu, Wenbin  and
      Zhang, Hanning  and
      Dong, Hanze  and
      Pi, Renjie  and
      Zhao, Han  and
      Jiang, Nan  and
      Ji, Heng  and
      Yao, Yuan  and
      Zhang, Tong",
    editor = "Al-Onaizan, Yaser  and
      Bansal, Mohit  and
      Chen, Yun-Nung",
    booktitle = "Proceedings of the 2024 Conference on Empirical Methods in Natural Language Processing",
    month = nov,
    year = "2024",
    address = "Miami, Florida, USA",
    publisher = "Association for Computational Linguistics",
    url = "https://aclanthology.org/2024.emnlp-main.35/",
    doi = "10.18653/v1/2024.emnlp-main.35",
    pages = "580--606"
}

@inproceedings{
zhai2023investigating,
title={Investigating the Catastrophic Forgetting in Multimodal Large Language Model Fine-Tuning},
author={Yuexiang Zhai and Shengbang Tong and Xiao Li and Mu Cai and Qing Qu and Yong Jae Lee and Yi Ma},
booktitle={Conference on Parsimony and Learning (Proceedings Track)},
year={2023},
url={https://openreview.net/forum?id=g7rMSiNtmA}
}

@article{zhao2025proactive,
  title={Proactive defense against LLM Jailbreak},
  author={Zhao, Weiliang and Peng, Jinjun and Ben-Levi, Daniel and Yu, Zhou and Yang, Junfeng},
  journal={arXiv preprint arXiv:2510.05052},
  year={2025}
}

@inproceedings{NEURIPS2024_094324f3,
 author = {Wang, Jiongxiao and Li, Jiazhao and Li, Yiquan and Qi, Xiangyu and Hu, Junjie and Li, Yixuan and McDaniel, Patrick and Chen, Muhao and Li, Bo and Xiao, Chaowei},
 booktitle = {Advances in Neural Information Processing Systems},
 doi = {10.52202/079017-0169},
 editor = {A. Globerson and L. Mackey and D. Belgrave and A. Fan and U. Paquet and J. Tomczak and C. Zhang},
 pages = {5210--5243},
 publisher = {Curran Associates, Inc.},
 title = {BackdoorAlign: Mitigating Fine-tuning based Jailbreak Attack with Backdoor Enhanced Safety Alignment},
 url = {https://proceedings.neurips.cc/paper_files/paper/2024/file/094324f386c836c75d4a26f3499d2ede-Paper-Conference.pdf},
 volume = {37},
 year = {2024}
}

@article{liu2023jailbreaking,
  title={Jailbreaking chatgpt via prompt engineering: An empirical study},
  author={Liu, Yi and Deng, Gelei and Xu, Zhengzi and Li, Yuekang and Zheng, Yaowen and Zhang, Ying and Zhao, Lida and Zhang, Tianwei and Wang, Kailong and Liu, Yang},
  journal={arXiv preprint arXiv:2305.13860},
  year={2023}
}

@inproceedings{shen2024anything,
  title={" do anything now": Characterizing and evaluating in-the-wild jailbreak prompts on large language models},
  author={Shen, Xinyue and Chen, Zeyuan and Backes, Michael and Shen, Yun and Zhang, Yang},
  booktitle={Proceedings of the 2024 on ACM SIGSAC Conference on Computer and Communications Security},
  pages={1671--1685},
  year={2024}
}

@inproceedings{ICLR2024_f83cb637,
 author = {Liu, Xiaogeng and Xu, Nan and Chen, Muhao and Xiao, Chaowei},
 booktitle = {International Conference on Representation Learning},
 editor = {B. Kim and Y. Yue and S. Chaudhuri and K. Fragkiadaki and M. Khan and Y. Sun},
 pages = {56174--56194},
 title = {AutoDAN: Generating Stealthy Jailbreak Prompts on Aligned Large Language Models},
 url = {https://proceedings.iclr.cc/paper_files/paper/2024/file/f83cb637e159e789f5576ff6848874de-Paper-Conference.pdf},
 volume = {2024},
 year = {2024}
}

@inproceedings{chao2025jailbreaking,
  title={Jailbreaking black box large language models in twenty queries},
  author={Chao, Patrick and Robey, Alexander and Dobriban, Edgar and Hassani, Hamed and Pappas, George J and Wong, Eric},
  booktitle={2025 IEEE Conference on Secure and Trustworthy Machine Learning (SaTML)},
  pages={23--42},
  year={2025},
  organization={IEEE}
}

@article{yu2023gptfuzzer,
  title={Gptfuzzer: Red teaming large language models with auto-generated jailbreak prompts},
  author={Yu, Jiahao and Lin, Xingwei and Yu, Zheng and Xing, Xinyu},
  journal={arXiv preprint arXiv:2309.10253},
  year={2023}
}

@misc{zhao2025armoraligningsecuresafe,
      title={ARMOR: Aligning Secure and Safe Large Language Models via Meticulous Reasoning}, 
      author={Zhengyue Zhao and Yingzi Ma and Somesh Jha and Marco Pavone and Patrick McDaniel and Chaowei Xiao},
      year={2025},
      eprint={2507.11500},
      archivePrefix={arXiv},
      primaryClass={cs.CR},
      url={https://arxiv.org/abs/2507.11500}, 
}

@inproceedings{
liu2024autodan,
title={Auto{DAN}: Generating Stealthy Jailbreak Prompts on Aligned Large Language Models},
author={Xiaogeng Liu and Nan Xu and Muhao Chen and Chaowei Xiao},
booktitle={The Twelfth International Conference on Learning Representations},
year={2024},
url={https://openreview.net/forum?id=7Jwpw4qKkb}
}

@article{cobbe2021training,
  title={Training verifiers to solve math word problems},
  author={Cobbe, Karl and Kosaraju, Vineet and Bavarian, Mohammad and Chen, Mark and Jun, Heewoo and Kaiser, Lukasz and Plappert, Matthias and Tworek, Jerry and Hilton, Jacob and Nakano, Reiichiro and others},
  journal={arXiv preprint arXiv:2110.14168},
  year={2021}
}

\end{document}